\pdfoutput=1

\documentclass[11pt]{article}
\usepackage{authblk}

\usepackage{acl}

\usepackage{times}
\usepackage{latexsym}

\usepackage[T1]{fontenc}

\usepackage[utf8]{inputenc}

\usepackage{tabularx}
\newcolumntype{Y}{>{\centering\arraybackslash}X}
\usepackage{booktabs}

\usepackage{microtype}
\usepackage[compact]{titlesec}

\usepackage{graphicx} 
\graphicspath{ {Figures/} }

\title{Two Contrasting Data Annotation Paradigms for Subjective NLP Tasks}

\author[1,2]{\textbf{Paul R\"ottger}}
\author[2]{\textbf{Bertie Vidgen}}
\author[3]{\textbf{Dirk Hovy}}
\author[1]{\textbf{Janet B. Pierrehumbert}}
\affil[1]{University of Oxford}
\affil[2]{The Alan Turing Institute}
\affil[3]{Bocconi University}

\begin{document}
\maketitle
\begin{abstract}
Labelled data is the foundation of most natural language processing tasks.
However, labelling data is difficult and there often are diverse valid beliefs about what the correct data labels should be.
So far, dataset creators have acknowledged annotator subjectivity, but rarely actively managed it in the annotation process.
This has led to partly-subjective datasets that fail to serve a clear downstream use.
To address this issue, we propose two contrasting paradigms for data annotation.
The \textit{descriptive} paradigm encourages annotator subjectivity, whereas the \textit{prescriptive} paradigm discourages it.
Descriptive annotation allows for the surveying and modelling of different beliefs, whereas prescriptive annotation enables the training of models that consistently apply one belief.
We discuss benefits and challenges in implementing both paradigms, and argue that dataset creators should explicitly aim for one or the other to facilitate the intended use of their dataset.
Lastly, we conduct an annotation experiment using hate speech data that illustrates the contrast between the two paradigms.
\end{abstract}

\section{Introduction}

Many natural language processing (NLP) tasks are \textit{subjective}, in the sense that there are diverse valid beliefs about what the correct data labels should be.
Some tasks, like hate speech detection, are highly subjective:
different people have very different beliefs about what should or should not be labelled as hateful \citep{waseem2016you, salminen2019hateratings, davani2021hate}, and while some beliefs are more widely accepted than others, there is no single objective truth.
Other examples include toxicity \citep{sap2019risk,sap2021annotators}, harassment \citep{alkuwatly2020identifying}, harmful content \citep{jiang2021understanding} and stance detection \citep{luo2020detecting, aldayel2021stance} as well as sentiment analysis \citep{kenyon2018sentiment,poria2020beneath}.
But even for seemingly objective tasks like part-of-speech tagging, there is subjective disagreement between annotators \citep{plank2014disagreement}.

\begin{figure}[t]
\centering
\includegraphics[width=0.48\textwidth]{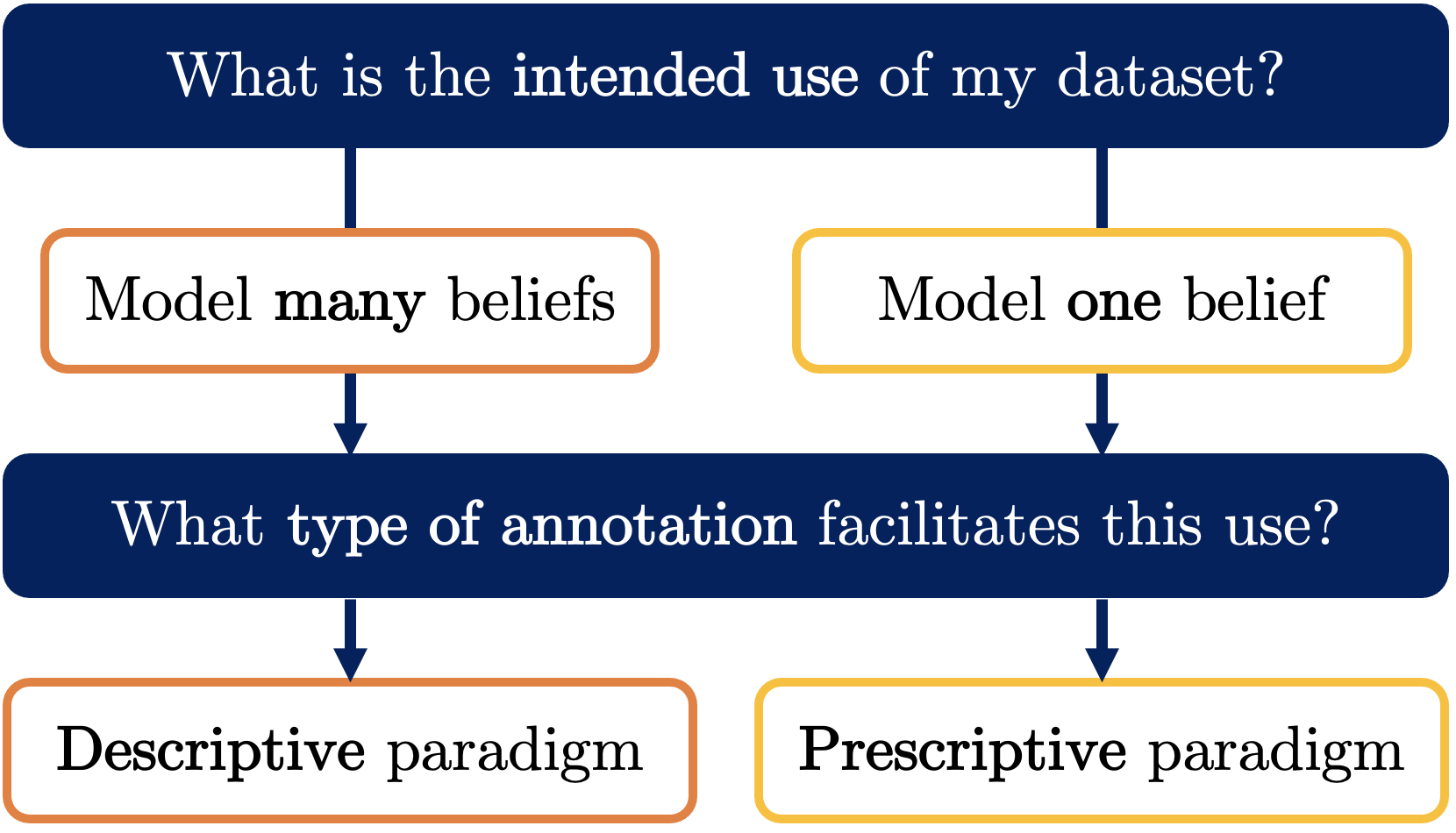}
\caption{Two key questions for dataset creators.}
\label{fig: schematic}
\end{figure}

In this article, we argue that dataset creators should consider the role of annotator subjectivity in the annotation process and either explicitly encourage it or discourage it.
Annotators may subjectively disagree about labels (e.g., for hate speech) but dataset creators can and should decide, based on the intended downstream use of their dataset, whether they want to a) capture \textit{different beliefs} or b) encode \textit{one specific belief} in their data.

As a framework, we propose two contrasting data annotation paradigms.
Each paradigm facilitates a clear and distinct downstream use.
The \textbf{descriptive} paradigm encourages annotator subjectivity to create datasets as granular surveys of individual beliefs.
Descriptive data annotation thus allows for the capturing and modelling of different beliefs.
The \textbf{prescriptive} paradigm, on the other hand, discourages annotator subjectivity and instead tasks annotators with encoding one specific belief, formulated in the annotation guidelines.
Prescriptive data annotation thus enables the training of models that seek to consistently apply one belief.
A researcher may, for example, want to model different beliefs about hate speech ($\rightarrow$ descriptive paradigm), while a content moderation engineer at a social media company may need models that apply their content policy ($\rightarrow$ prescriptive paradigm).

Neither paradigm is inherently superior, but explicitly aiming for one or the other is beneficial because it makes clear what an annotated dataset can and should be used for.
For example, data annotated under the descriptive paradigm can provide insights into different beliefs (\S\ref{subsec: descriptive - benefits}), but it cannot easily be used to train models with one pre-specified behaviour (\S\ref{subsec: prescriptive - benefits}).
By contrast, leaving annotator subjectivity unaddressed, as has mostly been the case in NLP so far, leads to datasets that neither capture an interpretable diversity of beliefs nor consistently encode one specific belief; an undesirable middle ground without a clear downstream use.\footnote{See Appx. \ref{app: datasets} for a selective overview of existing datasets.}

The two paradigms are applicable to all data annotation.
They can be used to compare existing datasets, and to make and communicate decisions about how new datasets are annotated as well as how annotator disagreement can be interpreted.
We hope that by naming and explaining the two paradigms, and by discussing key benefits and challenges in their implementation, we can support more intentional annotation process design, which will result in more useful NLP datasets.

\paragraph{Terminology}
Our use of the terms \textit{descriptive} and \textit{prescriptive} aligns with their use in both linguistics and ethics.
In linguistics, descriptivism studies how language \textit{is} used, whereas prescriptive grammar declares how language \textit{should} be used \citep{justice2006relevant}.
Descriptive ethics studies the moral judgments that people make, while prescriptive ethics considers how people ought to act \citep{thiroux2015ethics}.
Accordingly, descriptive data annotation surveys annotators' beliefs, whereas prescriptive data annotation aims to encode one specific belief, which is formulated in the annotation guidelines.


\section{The Descriptive Annotation Paradigm: Encouraging Annotator Subjectivity}

\subsection{Key Benefits} \label{subsec: descriptive - benefits}

\paragraph{Insights into Diverse Beliefs}
Descriptive data annotation captures a multiplicity of beliefs in data labels, much like a very granular survey would.
The distribution of data labels across annotators and examples can therefore provide insights into the beliefs of annotators, or the larger population they may represent.
For example, descriptive data annotation has shown that non-Black annotators are more likely to rate African American English as toxic \citep{sap2019risk,sap2021annotators}, and that people who identify as LGBTQ+ or young adults are more likely to rate random social media comments as toxic \citep{kumar2021designing}.
Similar correlations between sociodemographic characteristics and annotation outcomes have been found in stance \citep{luo2020detecting}, sentiment \citep{diaz2018addressing} and hate speech detection \citep{waseem2016you}.

Even very subjective tasks may have clear-cut entries on which most annotators agree.
For example, crowd workers tend to agree more on the extremes of a hate rating scale \citep{salminen2019hateratings}, and datasets which consist of clear hate and non-hate can have very high levels of inter-annotator agreement, even with minimal guidelines \citep{rottger2021hatecheck}.
Descriptive data annotation can help to identify which entries are more subjective.
\citet{jiang2021understanding}, for instance, find that perceptions about the harmfulness of sexually explicit language vary strongly across the eight countries in their sample, whereas support for mass murder or human trafficking is seen as very harmful across all countries.

\paragraph{Learning from Disagreement}
Annotator-level labels from descriptive data annotation have been shown to be a rich source of information for model training.
First, they can be used to separately model annotators' beliefs.
For less subjective tasks such as question answering, this has served to mitigate undesirable annotator biases \citep{geva2019modeling}.
\citet{davani2021disagreement} reframe and expand on this idea for more subjective tasks like abuse detection, showing that multi-annotator model architectures outperform standard single-label approaches on single label prediction.
Second, instead of modelling each annotator separately, other work has grouped them into clusters based on sociodemographic attributes \citep{alkuwatly2020identifying} or polarisation measures derived from annotator labels \citep{akhtar2020modeling,akhtar2021opinions}, with similar results.
Third, models can be trained directly on \textit{soft} labels (i.e., distributions of labels given by annotators), rather than \textit{hard} one-hot ground truth vectors \citep{plank2014learning,jamison2015softlabel,uma2020case,fornaciari2021softlabel}.

\paragraph{Evaluating with Disagreement}
Descriptive data annotation facilitates model evaluation that accounts for different beliefs about how a model should behave \citep{basile2021considerdisagreement, uma2021semevaldisagree}.
This is particularly relevant when deploying NLP systems for practical tasks such as content moderation, where \textit{user-facing} performance needs to be considered \citep{gordon2021disagreement}.
To this end, comparing a model prediction to a descriptive label distribution, the \textit{crowd truth} \citep{aroyo2015truth}, can help estimate how \textit{acceptable} the prediction would be to users \citep{alm2011subjective}.
\citet{gordon2022jury} operationalise this idea by introducing \textit{jury learning}, a recommender system approach to predicting how a group of annotators with specified sociodemographic characteristics would judge different pieces of content.

\subsection{Key Challenges} \label{subsec: descriptive - challenges}

\paragraph{Representativeness of Annotators}
The survey-like benefits of descriptive data annotation correspond to survey-like challenges.
First, dataset creators must decide who their data aims to represent, by establishing a clear population of interest.
\citet{arora2020harassment}, for example, ask women journalists to annotate harassment targeted at them.
\citet{waseem2016you} recruits feminist activists as well as crowd workers.
Second, dataset creators must consider whether representativeness can practically be achieved.
To capture a representative distribution of beliefs for each entry requires dozens, if not hundreds of annotators recruited from the population of interest.
\citet{sap2021annotators}, for example, collect toxicity labels from 641 annotators, but only for 15 examples.
Other datasets generally use much fewer annotators per entry (see Appx. \ref{app: datasets}) and therefore cannot be considered representative in the sense that large (i.e., many-participant) surveys are.
A potential approach to mitigating this issue in modelling annotator beliefs is by introducing information sharing across groups of annotators (e.g. based on sociodemographics), where annotator behaviour updates group-specific priors rather than being considered in isolation, and thus fewer annotations are needed from each annotator \citep{gordon2022jury}.

\paragraph{Interpretation of Disagreement}
In the descriptive paradigm, the absence of a (specified) ground truth label complicates the interpretation of any observed annotator disagreement:
it may be due to a genuine difference in beliefs, which is desirable in this paradigm, or due to undesirable annotator error \citep{pavlick2019inherent, basile2021toward, leonardelli2021agreeing}.
The same issue applies to inter-annotator agreement metrics like Fleiss' Kappa.
When subjectivity is encouraged, such metrics can at best measure task subjectiveness, but not task difficulty, annotator performance, or dataset quality \citep{zaenen2006markup,alm2011subjective}.

\paragraph{Label Aggregation}
Descriptive annotation has clear downstream uses (\S\ref{subsec: descriptive - benefits}) but it is fundamentally misaligned with standard NLP methods that rely on single gold standard labels.
When datasets are constructed to be granular surveys of beliefs, reducing those beliefs to a single label, through majority voting or otherwise, goes directly against that purpose.
Aggregating labels conceals informative disagreements \citep{leonardelli2021agreeing, basile2021considerdisagreement} and risks discarding minority beliefs \citep{prabhakaran2021releasing, basile2021toward}.

\section{The Prescriptive Annotation Paradigm: Discouraging Annotator Subjectivity}

\subsection{Key Benefits} \label{subsec: prescriptive - benefits}

\paragraph{Specified Model Behaviour}
Encoding one specific belief in a dataset through data annotation is difficult (\S\ref{subsec: prescriptive - challenges}) but advantageous for many practical applications.
Social media companies, for example, moderate content on their platforms according to specific and extensive content policies.\footnote{In March 2021, a whistleblower shared 300-page content guidelines used by Facebook moderators \citep{hern2021facebook}.}
Therefore, they need data annotated in accordance with those policies to train their content moderation models.
This illustrates that even for highly subjective tasks, where different model behaviours are plausible and valid, one specific behaviour may be practically desirable.
Prescriptive data annotation specifies such desired behaviours in datasets for model training and evaluation.

\paragraph{Quality Assurance}
In the prescriptive paradigm, annotator disagreements are a call to action because they indicate that a) the annotation guidelines were not correctly applied by annotators or b) the guidelines themselves were inadequate.
Annotator errors can be found using noise identification techniques \citep[e.g.,][]{hovy2013learning,zhang2017improving,paun2018comparingbayesian,northcutt2021confident}, corrected by expert annotators \citep{vidgen2020directions,vidgen2021contextual} or their impact mitigated by label aggregation.
Guidelines which are unclear or incomplete need to be clarified or expanded by dataset creators, which may require iterative approaches to annotation \citep{founta2018large,zeinert2021misogyny}.
Therefore, quality assurance under the prescriptive paradigm is a laborious but structured process, with inter-annotator agreement as a useful, albeit noisy, measure of dataset quality.

\paragraph{Visibility of Encoded Belief}
In the prescriptive paradigm, the one belief that annotators are tasked with applying is made visible and explicit in the annotation guidelines.
Well-formulated guidelines should give clear instructions on how to decide between different classes, along with explanations and illustrative examples.
This creates accountability, in that people can review, challenge and disagree with the formulated belief.
Like data statements \citep{bender2018data}, prescriptive annotation guidelines can provide detailed insights into how datasets were created, which can then inform their downstream use.

\subsection{Key Challenges} \label{subsec: prescriptive - challenges}

\paragraph{Creation of Annotation Guidelines}
Creating guidelines for prescriptive data annotation is difficult because it requires topical knowledge and familiarity with the data that is to be annotated.
Guidelines would ideally provide a clear judgment on every possible entry, but in practice, such perfectly comprehensive guidelines can only be approximated.
Even extensive legal definitions of hate speech leave some room for subjective interpretation \citep{sellars2016defining}.
Further, creating guidelines for prescriptive data annotation requires deciding which one belief to encode in the dataset.
This can be a complex process that risks disregarding non-majority beliefs if marginalised people are not included in it \citep{raji2020accountability}.

\paragraph{Application of Annotation Guidelines}
Annotators need to be familiar with annotation guidelines to apply them correctly, which may require additional training, especially if guidelines are long and complex.
This is reflected in an increasing shift in the literature towards using annotators with task-relevant experience over non-trained crowd workers \citep[e.g.][]{basile2019semeval,rottger2021hatecheck, vidgen2021contextual}.
During annotation, annotators will need to refer back to the guidelines, which requires giving them sufficient time per entry and providing a well-designed annotation interface.

\paragraph{Persistent Subjectivity}
Annotator subjectivity can be discouraged, but not eliminated.
Inevitable gaps in guidelines leave annotators no choice but to apply their personal judgement for some entries, and even when there is explicit guidance, implicit biases may persist.
\citet{sap2019risk}, for example, demonstrate racial biases in hate speech annotation, and show that targeted annotation prompts can reduce these biases but not definitively eliminate them.
To address this issue, dataset creators should work with groups of annotators that are diverse in terms of sociodemographic characteristics and personal experiences, even when annotator subjectivity is discouraged.

\section{An Illustrative Annotation Experiment} \label{sec: experiment}

\paragraph{Experimental Design}
To illustrate the contrast between the two paradigms, we conducted an annotation experiment.
60 annotators were randomly assigned to one of three groups of 20.
Each group was given different guidelines to label the same 200 Twitter posts, taken from a corpus annotated for hate speech by \citet{davidson2017automated}, as either \textit{hateful} or \textit{non-hateful}.
\textbf{G1}, the descriptive group, received a short prompt which directed them to apply their subjective judgement (`Do you personally feel this post is hateful?').
\textbf{G2}, the prescriptive group, received a short prompt which discouraged subjectivity (`Does this post meet the criteria for hate speech?'), along with detailed annotation guidelines.
\textbf{G3}, the control group, received the prescriptive prompt and a short definition of hate speech but no further guidelines.
This is to control for the difference in length and complexity of annotation guidelines between \textbf{G1} and \textbf{G2}.

\paragraph{Results}
We evaluate average percentage agreement and Fleiss' $\kappa$ to measure dataset-level inter-annotator agreement in each group (Table \ref{tab: experiment}).
To test for significant differences in agreement between groups, we use confidence intervals computed with a 1000-sample bootstrap.

\begin{table}[h]
\small
\centering
\begin{tabularx}{0.48\textwidth}{cYY}
\textbf{Group} & \textbf{Avg. \% Agree.} & \textbf{Fleiss' $\kappa$}\\
\toprule
\textbf{G1} - Descriptive & 73.90 & 0.20\\
\midrule
\textbf{G2} - Prescriptive & \textbf{93.72} & \textbf{0.78}\\
\midrule
\textbf{G3} - Control & 72.50 & 0.15\\
\bottomrule
\end{tabularx}
\caption{Inter-annotator agreement metrics for the three groups of 20 annotators on our 200-post binary dataset.}
\label{tab: experiment}
\end{table}

Agreement is very low in the descriptive group \textbf{G1} ($\kappa = 0.20$), which suggests that annotators hold varied beliefs about which posts are hateful.
However, agreement is significantly higher ($p<0.001$) in \textbf{G2} ($\kappa = 0.78$), which suggests that a prescriptive approach with detailed annotation guidelines can successfully induce annotators to apply a specified belief rather than their subjective view.
Further, agreement in the control group \textbf{G3} ($\kappa = 0.15$) is as low as in descriptive \textbf{G1}, which suggests that comprehensive guidelines are instrumental in facilitating high agreement in the prescriptive paradigm.
\textbf{G1} and \textbf{G3} also do not differ systematically on which posts annotators disagree on, which suggests that annotators with little prescriptive instruction (\textbf{G3}) tend to apply their subjective views (like \textbf{G1}).

\paragraph{Reproducibility}
For details on our dataset and annotators, see the data statement \citep{bender2018data} in Appendix \ref{app: data-statement}.
Annotation prompts are given in Appendix \ref{app: prompts}.
Full guidelines, annotated data and code are available \href{https://github.com/paul-rottger/annotation-paradigms}{on GitHub}.

\vspace{0.3cm}
\section{Conclusion} 

In this article, we named and explained two contrasting paradigms for data annotation.
The \textbf{descriptive} paradigm encourages annotator subjectivity to create datasets as granular surveys of individual beliefs, which can then be analysed and modelled.
The \textbf{prescriptive} paradigm tasks annotators with encoding one specific belief formulated in the annotation guidelines, to enable the training of models that seek to apply that one belief to unseen data.
Dataset creators should explicitly aim for one paradigm or the other to facilitate the intended downstream use of their dataset, and to document for the benefit of others how exactly their dataset was annotated.
We discussed benefits and challenges in implementing both paradigms, and conducted an annotation experiment that illustrates the contrast between them.
We hope that the two paradigms can support more intentional annotation process design and thus facilitate the creation of more useful NLP datasets.

\vspace{0.6cm}
\hrule
\vspace{0.1cm}
\hrule
\vspace{0.3cm}


\section*{Acknowledgments}
Paul Röttger was funded by the German Academic Scholarship Foundation.
Bertie Vidgen and Paul Röttger were both supported by The Alan Turing Institute and Towards Turing 2.0 under the EPSRC Grant EP/W037211/1.
Dirk Hovy received funding from the European Research Council (ERC) under the European Union’s Horizon 2020 research and innovation program (grant agreement No.\ 949944). He is a member of the Data and Marketing Insights (DMI) Unit of the Bocconi Institute for Data Science and Analysis (BIDSA).
Janet B. Pierrehumbert was supported by EPSRC Grant EP/T023333/1.
We thank the Milan NLP Group, the Groningen Computational Linguistics Group as well as the Pierrehumbert Language Modelling Group for helpful comments and all reviewers for their constructive feedback.

\bibliography{custom}

\ifdefined\DeclarePrefChars\DeclarePrefChars{'’-}\else\fi
\begin{thebibliography}{56}
\expandafter\ifx\csname natexlab\endcsname\relax\def\natexlab#1{#1}\fi

\bibitem[{Akhtar et~al.(2020)Akhtar, Basile, and Patti}]{akhtar2020modeling}
Sohail Akhtar, Valerio Basile, and Viviana Patti. 2020.
\newblock Modeling annotator perspective and polarized opinions to improve hate
  speech detection.
\newblock In \emph{Proceedings of the AAAI Conference on Human Computation and
  Crowdsourcing}, volume~8, pages 151--154.

\bibitem[{Akhtar et~al.(2021)Akhtar, Basile, and Patti}]{akhtar2021opinions}
Sohail Akhtar, Valerio Basile, and Viviana Patti. 2021.
\newblock Whose opinions matter? perspective-aware models to identify opinions
  of hate speech victims in abusive language detection.
\newblock \emph{arXiv preprint arXiv:2106.15896}.

\bibitem[{Al~Kuwatly et~al.(2020)Al~Kuwatly, Wich, and
  Groh}]{alkuwatly2020identifying}
Hala Al~Kuwatly, Maximilian Wich, and Georg Groh. 2020.
\newblock \href {https://doi.org/10.18653/v1/2020.alw-1.21} {Identifying and
  measuring annotator bias based on annotators{'} demographic characteristics}.
\newblock In \emph{Proceedings of the Fourth Workshop on Online Abuse and
  Harms}, pages 184--190, Online. Association for Computational Linguistics.

\bibitem[{AlDayel and Magdy(2021)}]{aldayel2021stance}
Abeer AlDayel and Walid Magdy. 2021.
\newblock Stance detection on social media: State of the art and trends.
\newblock \emph{Information Processing \& Management}, 58(4):102597.

\bibitem[{Alm(2011)}]{alm2011subjective}
Cecilia~Ovesdotter Alm. 2011.
\newblock \href {https://aclanthology.org/P11-2019} {Subjective natural
  language problems: Motivations, applications, characterizations, and
  implications}.
\newblock In \emph{Proceedings of the 49th Annual Meeting of the Association
  for Computational Linguistics: Human Language Technologies}, pages 107--112,
  Portland, Oregon, USA. Association for Computational Linguistics.

\bibitem[{Arora et~al.(2020)Arora, Guo, Levitan, McGregor, and
  Hirschberg}]{arora2020harassment}
Ishaan Arora, Julia Guo, Sarah~Ita Levitan, Susan McGregor, and Julia
  Hirschberg. 2020.
\newblock \href {https://doi.org/10.18653/v1/2020.alw-1.2} {A novel methodology
  for developing automatic harassment classifiers for {T}witter}.
\newblock In \emph{Proceedings of the Fourth Workshop on Online Abuse and
  Harms}, pages 7--15, Online. Association for Computational Linguistics.

\bibitem[{Aroyo and Welty(2015)}]{aroyo2015truth}
Lora Aroyo and Chris Welty. 2015.
\newblock Truth is a lie: Crowd truth and the seven myths of human annotation.
\newblock \emph{AI Magazine}, 36(1):15--24.

\bibitem[{Basile et~al.(2019)Basile, Bosco, Fersini, Nozza, Patti, Pardo,
  Rosso, and Sanguinetti}]{basile2019semeval}
Valerio Basile, Cristina Bosco, Elisabetta Fersini, Debora Nozza, Viviana
  Patti, Francisco Manuel~Rangel Pardo, Paolo Rosso, and Manuela Sanguinetti.
  2019.
\newblock Semeval-2019 task 5: Multilingual detection of hate speech against
  immigrants and women in {T}witter.
\newblock In \emph{Proceedings of the 13th International Workshop on Semantic
  Evaluation}, pages 54--63.

\bibitem[{Basile et~al.(2021{\natexlab{a}})Basile, Cabitza, Campagner, and
  Fell}]{basile2021toward}
Valerio Basile, Federico Cabitza, Andrea Campagner, and Michael Fell.
  2021{\natexlab{a}}.
\newblock Toward a perspectivist turn in ground truthing for predictive
  computing.
\newblock \emph{Conference of the Italian Chapter of the Association for
  Intelligent Systems (ItAIS 2021)}.

\bibitem[{Basile et~al.(2021{\natexlab{b}})Basile, Fell, Fornaciari, Hovy,
  Paun, Plank, Poesio, and Uma}]{basile2021considerdisagreement}
Valerio Basile, Michael Fell, Tommaso Fornaciari, Dirk Hovy, Silviu Paun,
  Barbara Plank, Massimo Poesio, and Alexandra Uma. 2021{\natexlab{b}}.
\newblock \href {https://doi.org/10.18653/v1/2021.bppf-1.3} {We need to
  consider disagreement in evaluation}.
\newblock In \emph{Proceedings of the 1st Workshop on Benchmarking: Past,
  Present and Future}, pages 15--21, Online. Association for Computational
  Linguistics.

\bibitem[{Bender and Friedman(2018)}]{bender2018data}
Emily~M. Bender and Batya Friedman. 2018.
\newblock \href {https://doi.org/10.1162/tacl_a_00041} {Data statements for
  natural language processing: Toward mitigating system bias and enabling
  better science}.
\newblock \emph{Transactions of the Association for Computational Linguistics},
  6:587--604.

\bibitem[{Caselli et~al.(2020)Caselli, Basile, Mitrovi{\'c}, Kartoziya, and
  Granitzer}]{caselli2020feeloffended}
Tommaso Caselli, Valerio Basile, Jelena Mitrovi{\'c}, Inga Kartoziya, and
  Michael Granitzer. 2020.
\newblock \href {https://aclanthology.org/2020.lrec-1.760} {{I} feel offended,
  don{'}t be abusive! implicit/explicit messages in offensive and abusive
  language}.
\newblock In \emph{Proceedings of the 12th Language Resources and Evaluation
  Conference}, pages 6193--6202, Marseille, France. European Language Resources
  Association.

\bibitem[{Cercas~Curry et~al.(2021)Cercas~Curry, Abercrombie, and
  Rieser}]{cercas2021convabuse}
Amanda Cercas~Curry, Gavin Abercrombie, and Verena Rieser. 2021.
\newblock \href {https://aclanthology.org/2021.emnlp-main.587} {{C}onv{A}buse:
  Data, analysis, and benchmarks for nuanced detection in conversational {AI}}.
\newblock In \emph{Proceedings of the 2021 Conference on Empirical Methods in
  Natural Language Processing}, pages 7388--7403, Online and Punta Cana,
  Dominican Republic. Association for Computational Linguistics.

\bibitem[{Davani et~al.(2021{\natexlab{a}})Davani, Atari, Kennedy, and
  Dehghani}]{davani2021hate}
Aida~Mostafazadeh Davani, Mohammad Atari, Brendan Kennedy, and Morteza
  Dehghani. 2021{\natexlab{a}}.
\newblock \href {http://arxiv.org/abs/2110.14839} {Hate speech classifiers
  learn human-like social stereotypes}.

\bibitem[{Davani et~al.(2021{\natexlab{b}})Davani, D{\'\i}az, and
  Prabhakaran}]{davani2021disagreement}
Aida~Mostafazadeh Davani, Mark D{\'\i}az, and Vinodkumar Prabhakaran.
  2021{\natexlab{b}}.
\newblock Dealing with disagreements: Looking beyond the majority vote in
  subjective annotations.
\newblock \emph{arXiv preprint arXiv:2110.05719}.

\bibitem[{Davidson et~al.(2017)Davidson, Warmsley, Macy, and
  Weber}]{davidson2017automated}
Thomas Davidson, Dana Warmsley, Michael Macy, and Ingmar Weber. 2017.
\newblock Automated hate speech detection and the problem of offensive
  language.
\newblock In \emph{Proceedings of the 11th International AAAI Conference on Web
  and Social Media}, pages 512--515. Association for the Advancement of
  Artificial Intelligence.

\bibitem[{Diaz et~al.(2018)Diaz, Johnson, Lazar, Piper, and
  Gergle}]{diaz2018addressing}
Mark Diaz, Isaac Johnson, Amanda Lazar, Anne~Marie Piper, and Darren Gergle.
  2018.
\newblock \href {https://doi.org/10.1145/3173574.3173986} {\emph{Addressing
  Age-Related Bias in Sentiment Analysis}}, page 1–14. Association for
  Computing Machinery, New York, NY, USA.

\bibitem[{Fornaciari et~al.(2021)Fornaciari, Uma, Paun, Plank, Hovy, and
  Poesio}]{fornaciari2021softlabel}
Tommaso Fornaciari, Alexandra Uma, Silviu Paun, Barbara Plank, Dirk Hovy, and
  Massimo Poesio. 2021.
\newblock \href {https://doi.org/10.18653/v1/2021.naacl-main.204} {Beyond black
  {\&} white: Leveraging annotator disagreement via soft-label multi-task
  learning}.
\newblock In \emph{Proceedings of the 2021 Conference of the North American
  Chapter of the Association for Computational Linguistics: Human Language
  Technologies}, pages 2591--2597, Online. Association for Computational
  Linguistics.

\bibitem[{Founta et~al.(2018)Founta, Djouvas, Chatzakou, Leontiadis, Blackburn,
  Stringhini, Vakali, Sirivianos, and Kourtellis}]{founta2018large}
Antigoni~Maria Founta, Constantinos Djouvas, Despoina Chatzakou, Ilias
  Leontiadis, Jeremy Blackburn, Gianluca Stringhini, Athena Vakali, Michael
  Sirivianos, and Nicolas Kourtellis. 2018.
\newblock Large scale crowdsourcing and characterization of {T}witter abusive
  behavior.
\newblock In \emph{Proceedings of the 12th International AAAI Conference on Web
  and Social Media}, pages 491--500. Association for the Advancement of
  Artificial Intelligence.

\bibitem[{Geva et~al.(2019)Geva, Goldberg, and Berant}]{geva2019modeling}
Mor Geva, Yoav Goldberg, and Jonathan Berant. 2019.
\newblock \href {https://doi.org/10.18653/v1/D19-1107} {Are we modeling the
  task or the annotator? an investigation of annotator bias in natural language
  understanding datasets}.
\newblock In \emph{Proceedings of the 2019 Conference on Empirical Methods in
  Natural Language Processing and the 9th International Joint Conference on
  Natural Language Processing (EMNLP-IJCNLP)}, pages 1161--1166, Hong Kong,
  China. Association for Computational Linguistics.

\bibitem[{Gordon et~al.(2022)Gordon, Lam, Park, Patel, Hancock, Hashimoto, and
  Bernstein}]{gordon2022jury}
Mitchell~L Gordon, Michelle~S Lam, Joon~Sung Park, Kayur Patel, Jeffrey~T
  Hancock, Tatsunori Hashimoto, and Michael~S Bernstein. 2022.
\newblock Jury learning: Integrating dissenting voices into machine learning
  models.
\newblock \emph{arXiv preprint arXiv:2202.02950}.

\bibitem[{Gordon et~al.(2021)Gordon, Zhou, Patel, Hashimoto, and
  Bernstein}]{gordon2021disagreement}
Mitchell~L Gordon, Kaitlyn Zhou, Kayur Patel, Tatsunori Hashimoto, and
  Michael~S Bernstein. 2021.
\newblock The disagreement deconvolution: Bringing machine learning performance
  metrics in line with reality.
\newblock In \emph{Proceedings of the 2021 CHI Conference on Human Factors in
  Computing Systems}, pages 1--14.

\bibitem[{Hern(2021)}]{hern2021facebook}
Alex Hern. 2021.
\newblock Decoding emojis and defining 'support': Facebook's rules for content
  revealed.
\newblock \emph{The Guardian}.

\bibitem[{Hovy et~al.(2013)Hovy, Berg-Kirkpatrick, Vaswani, and
  Hovy}]{hovy2013learning}
Dirk Hovy, Taylor Berg-Kirkpatrick, Ashish Vaswani, and Eduard Hovy. 2013.
\newblock \href {https://aclanthology.org/N13-1132} {Learning whom to trust
  with {MACE}}.
\newblock In \emph{Proceedings of the 2013 Conference of the North {A}merican
  Chapter of the Association for Computational Linguistics: Human Language
  Technologies}, pages 1120--1130, Atlanta, Georgia. Association for
  Computational Linguistics.

\bibitem[{Jamison and Gurevych(2015)}]{jamison2015softlabel}
Emily Jamison and Iryna Gurevych. 2015.
\newblock \href {https://doi.org/10.18653/v1/D15-1035} {Noise or additional
  information? leveraging crowdsource annotation item agreement for natural
  language tasks.}
\newblock In \emph{Proceedings of the 2015 Conference on Empirical Methods in
  Natural Language Processing}, pages 291--297, Lisbon, Portugal. Association
  for Computational Linguistics.

\bibitem[{Jiang et~al.(2021)Jiang, Scheuerman, Fiesler, and
  Brubaker}]{jiang2021understanding}
Jialun~Aaron Jiang, Morgan~Klaus Scheuerman, Casey Fiesler, and Jed~R Brubaker.
  2021.
\newblock Understanding international perceptions of the severity of harmful
  content online.
\newblock \emph{PloS one}, 16(8):e0256762.

\bibitem[{Justice(2006)}]{justice2006relevant}
Paul Justice. 2006.
\newblock \emph{Relevant Linguistics}, 2nd edition.
\newblock Center for the Study of Language and Information, Stanford
  University.

\bibitem[{Kenyon-Dean et~al.(2018)Kenyon-Dean, Ahmed, Fujimoto,
  Georges-Filteau, Glasz, Kaur, Lalande, Bhanderi, Belfer, Kanagasabai,
  Sarrazingendron, Verma, and Ruths}]{kenyon2018sentiment}
Kian Kenyon-Dean, Eisha Ahmed, Scott Fujimoto, Jeremy Georges-Filteau,
  Christopher Glasz, Barleen Kaur, Auguste Lalande, Shruti Bhanderi, Robert
  Belfer, Nirmal Kanagasabai, Roman Sarrazingendron, Rohit Verma, and Derek
  Ruths. 2018.
\newblock \href {https://doi.org/10.18653/v1/N18-1171} {Sentiment analysis:
  It{'}s complicated!}
\newblock In \emph{Proceedings of the 2018 Conference of the North {A}merican
  Chapter of the Association for Computational Linguistics: Human Language
  Technologies, Volume 1 (Long Papers)}, pages 1886--1895, New Orleans,
  Louisiana. Association for Computational Linguistics.

\bibitem[{Kumar et~al.(2021)Kumar, Kelley, Consolvo, Mason, Bursztein,
  Durumeric, Thomas, and Bailey}]{kumar2021designing}
Deepak Kumar, Patrick~Gage Kelley, Sunny Consolvo, Joshua Mason, Elie
  Bursztein, Zakir Durumeric, Kurt Thomas, and Michael Bailey. 2021.
\newblock Designing toxic content classification for a diversity of
  perspectives.
\newblock In \emph{Seventeenth Symposium on Usable Privacy and Security (SOUPS
  2021)}, pages 299--318.

\bibitem[{Leonardelli et~al.(2021)Leonardelli, Menini, Palmero~Aprosio,
  Guerini, and Tonelli}]{leonardelli2021agreeing}
Elisa Leonardelli, Stefano Menini, Alessio Palmero~Aprosio, Marco Guerini, and
  Sara Tonelli. 2021.
\newblock \href {https://aclanthology.org/2021.emnlp-main.822} {Agreeing to
  disagree: Annotating offensive language datasets with annotators{'}
  disagreement}.
\newblock In \emph{Proceedings of the 2021 Conference on Empirical Methods in
  Natural Language Processing}, pages 10528--10539, Online and Punta Cana,
  Dominican Republic. Association for Computational Linguistics.

\bibitem[{Luo et~al.(2020)Luo, Card, and Jurafsky}]{luo2020detecting}
Yiwei Luo, Dallas Card, and Dan Jurafsky. 2020.
\newblock \href {https://doi.org/10.18653/v1/2020.findings-emnlp.296}
  {Detecting stance in media on global warming}.
\newblock In \emph{Findings of the Association for Computational Linguistics:
  EMNLP 2020}, pages 3296--3315, Online. Association for Computational
  Linguistics.

\bibitem[{Northcutt et~al.(2021)Northcutt, Jiang, and
  Chuang}]{northcutt2021confident}
Curtis Northcutt, Lu~Jiang, and Isaac Chuang. 2021.
\newblock Confident learning: Estimating uncertainty in dataset labels.
\newblock \emph{Journal of Artificial Intelligence Research}, 70:1373--1411.

\bibitem[{Paun et~al.(2018)Paun, Carpenter, Chamberlain, Hovy, Kruschwitz, and
  Poesio}]{paun2018comparingbayesian}
Silviu Paun, Bob Carpenter, Jon Chamberlain, Dirk Hovy, Udo Kruschwitz, and
  Massimo Poesio. 2018.
\newblock \href {https://doi.org/10.1162/tacl_a_00040} {Comparing {B}ayesian
  models of annotation}.
\newblock \emph{Transactions of the Association for Computational Linguistics},
  6:571--585.

\bibitem[{Pavlick and Kwiatkowski(2019)}]{pavlick2019inherent}
Ellie Pavlick and Tom Kwiatkowski. 2019.
\newblock \href {https://doi.org/10.1162/tacl_a_00293} {Inherent disagreements
  in human textual inferences}.
\newblock \emph{Transactions of the Association for Computational Linguistics},
  7:677--694.

\bibitem[{Plank et~al.(2014{\natexlab{a}})Plank, Hovy, and
  S{\o}gaard}]{plank2014learning}
Barbara Plank, Dirk Hovy, and Anders S{\o}gaard. 2014{\natexlab{a}}.
\newblock \href {https://doi.org/10.3115/v1/E14-1078} {Learning part-of-speech
  taggers with inter-annotator agreement loss}.
\newblock In \emph{Proceedings of the 14th Conference of the {E}uropean Chapter
  of the Association for Computational Linguistics}, pages 742--751,
  Gothenburg, Sweden. Association for Computational Linguistics.

\bibitem[{Plank et~al.(2014{\natexlab{b}})Plank, Hovy, and
  S{\o}gaard}]{plank2014disagreement}
Barbara Plank, Dirk Hovy, and Anders S{\o}gaard. 2014{\natexlab{b}}.
\newblock \href {https://doi.org/10.3115/v1/P14-2083} {Linguistically debatable
  or just plain wrong?}
\newblock In \emph{Proceedings of the 52nd Annual Meeting of the Association
  for Computational Linguistics (Volume 2: Short Papers)}, pages 507--511,
  Baltimore, Maryland. Association for Computational Linguistics.

\bibitem[{Poria et~al.(2020)Poria, Hazarika, Majumder, and
  Mihalcea}]{poria2020beneath}
Soujanya Poria, Devamanyu Hazarika, Navonil Majumder, and Rada Mihalcea. 2020.
\newblock Beneath the tip of the iceberg: Current challenges and new directions
  in sentiment analysis research.
\newblock \emph{IEEE Transactions on Affective Computing}.

\bibitem[{Prabhakaran et~al.(2021)Prabhakaran, Mostafazadeh~Davani, and
  Diaz}]{prabhakaran2021releasing}
Vinodkumar Prabhakaran, Aida Mostafazadeh~Davani, and Mark Diaz. 2021.
\newblock \href {https://aclanthology.org/2021.law-1.14} {On releasing
  annotator-level labels and information in datasets}.
\newblock In \emph{Proceedings of The Joint 15th Linguistic Annotation Workshop
  (LAW) and 3rd Designing Meaning Representations (DMR) Workshop}, pages
  133--138, Punta Cana, Dominican Republic. Association for Computational
  Linguistics.

\bibitem[{Raji et~al.(2020)Raji, Smart, White, Mitchell, Gebru, Hutchinson,
  Smith-Loud, Theron, and Barnes}]{raji2020accountability}
Inioluwa~Deborah Raji, Andrew Smart, Rebecca~N. White, Margaret Mitchell,
  Timnit Gebru, Ben Hutchinson, Jamila Smith-Loud, Daniel Theron, and Parker
  Barnes. 2020.
\newblock \href {https://doi.org/10.1145/3351095.3372873} {Closing the ai
  accountability gap: Defining an end-to-end framework for internal algorithmic
  auditing}.
\newblock In \emph{Proceedings of the 2020 Conference on Fairness,
  Accountability, and Transparency}, FAT* '20, page 33–44, New York, NY, USA.
  Association for Computing Machinery.

\bibitem[{R{\"o}ttger et~al.(2021)R{\"o}ttger, Vidgen, Nguyen, Talat, Margetts,
  and Pierrehumbert}]{rottger2021hatecheck}
Paul R{\"o}ttger, Bertie Vidgen, Dong Nguyen, Zeerak Talat, Helen Margetts, and
  Janet Pierrehumbert. 2021.
\newblock \href {https://doi.org/10.18653/v1/2021.acl-long.4} {{H}ate{C}heck:
  Functional tests for hate speech detection models}.
\newblock In \emph{Proceedings of the 59th Annual Meeting of the Association
  for Computational Linguistics and the 11th International Joint Conference on
  Natural Language Processing (Volume 1: Long Papers)}, pages 41--58, Online.
  Association for Computational Linguistics.

\bibitem[{Salminen et~al.(2019)Salminen, Almerekhi, Kamel, Jung, and
  Jansen}]{salminen2019hateratings}
Joni Salminen, Hind Almerekhi, Ahmed~Mohamed Kamel, Soon-gyo Jung, and
  Bernard~J. Jansen. 2019.
\newblock \href {https://doi.org/10.1145/3295750.3298954} {Online hate ratings
  vary by extremes: A statistical analysis}.
\newblock In \emph{Proceedings of the 2019 Conference on Human Information
  Interaction and Retrieval}, CHIIR '19, page 213–217, New York, NY, USA.
  Association for Computing Machinery.

\bibitem[{Sap et~al.(2019)Sap, Card, Gabriel, Choi, and Smith}]{sap2019risk}
Maarten Sap, Dallas Card, Saadia Gabriel, Yejin Choi, and Noah~A. Smith. 2019.
\newblock \href {https://doi.org/10.18653/v1/P19-1163} {The risk of racial bias
  in hate speech detection}.
\newblock In \emph{Proceedings of the 57th Annual Meeting of the Association
  for Computational Linguistics}, pages 1668--1678, Florence, Italy.
  Association for Computational Linguistics.

\bibitem[{Sap et~al.(2021)Sap, Swayamdipta, Vianna, Zhou, Choi, and
  Smith}]{sap2021annotators}
Maarten Sap, Swabha Swayamdipta, Laura Vianna, Xuhui Zhou, Yejin Choi, and
  Noah~A. Smith. 2021.
\newblock \href {http://arxiv.org/abs/2111.07997} {Annotators with attitudes:
  How annotator beliefs and identities bias toxic language detection}.

\bibitem[{Sellars(2016)}]{sellars2016defining}
Andrew Sellars. 2016.
\newblock Defining hate speech.
\newblock \emph{Berkman Klein Center Research Publication}, 20(2016):16--48.

\bibitem[{Talat(2016)}]{waseem2016you}
Zeerak Talat. 2016.
\newblock \href {https://doi.org/10.18653/v1/W16-5618} {Are you a racist or am
  {I} seeing things? {A}nnotator influence on hate speech detection on
  {T}witter}.
\newblock In \emph{Proceedings of the First Workshop on {NLP} and Computational
  Social Science}, pages 138--142, Austin, Texas. Association for Computational
  Linguistics.

\bibitem[{Talat and Hovy(2016)}]{waseem2016hateful}
Zeerak Talat and Dirk Hovy. 2016.
\newblock \href {https://doi.org/10.18653/v1/N16-2013} {Hateful symbols or
  hateful people? {P}redictive features for hate speech detection on
  {T}witter}.
\newblock In \emph{Proceedings of the {NAACL} Student Research Workshop}, pages
  88--93, San Diego, California. Association for Computational Linguistics.

\bibitem[{Thiroux and Krasemann(2015)}]{thiroux2015ethics}
Jacques~P Thiroux and Keith~W Krasemann. 2015.
\newblock \emph{Ethics: Theory and Practice}, 11th edition.
\newblock Pearson.

\bibitem[{Uma et~al.(2021)Uma, Fornaciari, Dumitrache, Miller, Chamberlain,
  Plank, Simpson, and Poesio}]{uma2021semevaldisagree}
Alexandra Uma, Tommaso Fornaciari, Anca Dumitrache, Tristan Miller, Jon
  Chamberlain, Barbara Plank, Edwin Simpson, and Massimo Poesio. 2021.
\newblock \href {https://doi.org/10.18653/v1/2021.semeval-1.41}
  {{S}em{E}val-2021 task 12: Learning with disagreements}.
\newblock In \emph{Proceedings of the 15th International Workshop on Semantic
  Evaluation (SemEval-2021)}, pages 338--347, Online. Association for
  Computational Linguistics.

\bibitem[{Uma et~al.(2020)Uma, Fornaciari, Hovy, Paun, Plank, and
  Poesio}]{uma2020case}
Alexandra Uma, Tommaso Fornaciari, Dirk Hovy, Silviu Paun, Barbara Plank, and
  Massimo Poesio. 2020.
\newblock A case for soft loss functions.
\newblock \emph{Proceedings of the AAAI Conference on Human Computation and
  Crowdsourcing}, 8(1):173--177.

\bibitem[{Vidgen and Derczynski(2020)}]{vidgen2020directions}
Bertie Vidgen and Leon Derczynski. 2020.
\newblock \href {https://doi.org/10.1371/journal.pone.0243300} {Directions in
  abusive language training data, a systematic review: {{Garbage}} in, garbage
  out}.
\newblock \emph{PLOS ONE}, 15(12):e0243300.

\bibitem[{Vidgen et~al.(2021{\natexlab{a}})Vidgen, Nguyen, Margetts, Rossini,
  and Tromble}]{vidgen2021contextual}
Bertie Vidgen, Dong Nguyen, Helen Margetts, Patricia Rossini, and Rebekah
  Tromble. 2021{\natexlab{a}}.
\newblock \href {https://doi.org/10.18653/v1/2021.naacl-main.182} {Introducing
  {CAD}: the contextual abuse dataset}.
\newblock In \emph{Proceedings of the 2021 Conference of the North American
  Chapter of the Association for Computational Linguistics: Human Language
  Technologies}, pages 2289--2303, Online. Association for Computational
  Linguistics.

\bibitem[{Vidgen et~al.(2021{\natexlab{b}})Vidgen, Thrush, Talat, and
  Kiela}]{vidgen2021learning}
Bertie Vidgen, Tristan Thrush, Zeerak Talat, and Douwe Kiela.
  2021{\natexlab{b}}.
\newblock \href {https://doi.org/10.18653/v1/2021.acl-long.132} {Learning from
  the worst: Dynamically generated datasets to improve online hate detection}.
\newblock In \emph{Proceedings of the 59th Annual Meeting of the Association
  for Computational Linguistics and the 11th International Joint Conference on
  Natural Language Processing (Volume 1: Long Papers)}, pages 1667--1682,
  Online. Association for Computational Linguistics.

\bibitem[{Zaenen(2006)}]{zaenen2006markup}
Annie Zaenen. 2006.
\newblock \href {https://doi.org/10.1162/coli.2006.32.4.577} {Mark-up barking
  up the wrong tree}.
\newblock \emph{Comput. Linguist.}, 32(4):577–580.

\bibitem[{Zampieri et~al.(2019)Zampieri, Malmasi, Nakov, Rosenthal, Farra, and
  Kumar}]{zampieri2019predicting}
Marcos Zampieri, Shervin Malmasi, Preslav Nakov, Sara Rosenthal, Noura Farra,
  and Ritesh Kumar. 2019.
\newblock \href {https://doi.org/10.18653/v1/N19-1144} {Predicting the type and
  target of offensive posts in social media}.
\newblock In \emph{Proceedings of the 2019 Conference of the North {A}merican
  Chapter of the Association for Computational Linguistics: Human Language
  Technologies, Volume 1 (Long and Short Papers)}, pages 1415--1420,
  Minneapolis, Minnesota. Association for Computational Linguistics.

\bibitem[{Zeinert et~al.(2021)Zeinert, Inie, and
  Derczynski}]{zeinert2021misogyny}
Philine Zeinert, Nanna Inie, and Leon Derczynski. 2021.
\newblock \href {https://doi.org/10.18653/v1/2021.acl-long.247} {Annotating
  online misogyny}.
\newblock In \emph{Proceedings of the 59th Annual Meeting of the Association
  for Computational Linguistics and the 11th International Joint Conference on
  Natural Language Processing (Volume 1: Long Papers)}, pages 3181--3197,
  Online. Association for Computational Linguistics.

\bibitem[{Zhang et~al.(2017)Zhang, Sheng, Li, and Wu}]{zhang2017improving}
Jing Zhang, Victor~S Sheng, Tao Li, and Xindong Wu. 2017.
\newblock Improving crowdsourced label quality using noise correction.
\newblock \emph{IEEE transactions on neural networks and learning systems},
  29(5):1675--1688.

\end{thebibliography}
\bibliographystyle{acl_natbib}
\clearpage

\appendix

\section{Overview of Subjective Task Datasets} \label{app: datasets}

This appendix gives a selective overview of how existing NLP dataset work has (or has not) engaged with annotator subjectivity.
For reasons of scope, we focus on 11 English-language datasets annotated for hate speech and other forms of abuse.
Entries are sorted from most descriptive to most prescriptive annotation, based on our assessment of information made available by the dataset creators.

\citet{sap2019risk} and \citet{sap2021annotators} annotate toxicity.
They do not state explicitly that they encourage annotator subjectivity, but their annotation prompts clearly do.
Each entry is labelled by up to 641 annotators.
Overall, they are \textbf{very aligned with the descriptive paradigm}.

\citet{kumar2021designing} annotate toxicity and types of toxicity.
They do not state explicitly that they encourage annotator subjectivity, but their annotation prompts clearly do.
Each entry is labelled by five annotators.
Overall, they are \textbf{very aligned with the descriptive paradigm}.

\citet{cercas2021convabuse} annotate abuse.
They gather `views of expert annotators' based on guidelines that allow for significant subjectivity and do not attempt to resolve disagreements, but also do not explicitly encourage annotator subjectivity.
On average, each entry is labelled by around three annotators.
Overall, they are \textbf{moderately aligned with the descriptive paradigm}.

\citet{waseem2016hateful} annotate hate speech.
They provide annotators with 11 fine-grained criteria for hate speech, but several criteria invite subjective responses (e.g., `uses a \textit{problematic} hashtag').
Each entry is labelled by up to three annotators.
Overall, they are \textbf{not clearly aligned with either paradigm}.

\citet{davidson2017automated} annotate hate speech.
They provide annotators with a brief definition of hate speech and an explanatory paragraph, but their definition also includes subjective criteria like perceived `intent'.
Most entries are labelled by three annotators.
Overall, they are \textbf{not clearly aligned with either paradigm}.

\citet{zampieri2019predicting} annotate offensive content.
They provide annotators with some formal criteria for offensiveness (e.g., `use of profanity'), but as a whole their guidelines are very brief.
Each entry is labelled by up to three annotators.
Overall, they are \textbf{moderately aligned with the prescriptive paradigm}.

\citet{founta2018large} annotate abuse.
They provide annotators with fine-grained definitions for each category and iterate on their taxonomy to facilitate more agreement, but do not share comprehensive guidelines.
Each entry is labelled by five annotators.
Overall, they are \textbf{moderately aligned with the prescriptive paradigm}.

\citet{caselli2020feeloffended} annotate abuse.
They provide annotators with a brief fine-grained decision tree with the explicit intent of reducing annotator subjectivity, and discuss disagreements to resolve them.
Each entry is labelled by up to three annotators.
Overall, they are \textbf{moderately aligned with the prescriptive paradigm}.

\citet{vidgen2021learning} annotate hate speech.
They provide annotators with fine-grained definitions for each category as well as very detailed annotation guidelines, and disagreements are resolved by an expert.
Each entry is labelled by up to three annotators.
Overall, they are \textbf{very aligned with the prescriptive paradigm}.

\citet{vidgen2021contextual} annotate abuse.
They provide annotators with fine-grained definitions for each category as well as very detailed annotation guidelines, and they use expert-driven group adjudication to resolve disagreements.
Each entry is labelled by up to three annotators.
Overall, they are \textbf{very aligned with the prescriptive paradigm}.\\

\section{Data Statement} \label{app: data-statement}
Following \citet{bender2018data}, we provide a data statement, which documents the generation and provenance of the dataset used for our annotation experiment.

\paragraph{A. CURATION RATIONALE}
To create our dataset, we sampled 200 Twitter posts from a larger corpus annotated for hateful content by \citet{davidson2017automated}.
Of the posts we sampled, 100 were originally annotated as hateful and 100 as non-hateful by majority vote between three annotators.
We sampled only from those posts that had some disagreement among their annotators (i.e., two out of three rather than unanimous agreement), to encourage disagreement in our experiment.
The purpose of our 200-post dataset is to enable the annotation experiment presented in \S\ref{tab: experiment}, which illustrates the contrast between the descriptive and prescriptive data annotation paradigms.

\paragraph{B. LANGUAGE VARIETY}
The dataset contains English-language text posts only.

\paragraph{C. SPEAKER DEMOGRAPHICS}
All speakers are Twitter users.
\citet{davidson2017automated} do not share any information on their demographics.

\paragraph{D. ANNOTATOR RECRUITMENT}
We recruited three groups of 20 annotators using Amazon's Mechanical Turk crowdsourcing marketplace.\footnote{\href{www.mturk.com}{https://www.mturk.com/}}.
Annotators were made aware that the task contained instances of offensive language before starting their work, and they could withdraw at any point throughout the work.

\paragraph{E. ANNOTATOR DEMOGRAPHICS}
All annotators were at least 18 years old when they started their work, and we recruited only annotators that were based in the UK.
This was to facilitate comparability across groups of annotators.
For each group, we recruited 10 male and 10 female annotators, based on self-reported gender.
This was to encourage disagreement within groups, based on the assumption that men would on average disagree more about hateful content with women than with other men, and vice versa. No further annotator demographics were recorded.

\paragraph{F. ANNOTATOR COMPENSATION}
All annotators were compensated for their work at a rate of at least £16 per hour. The rate was set 50\% above the London living wage (£10.85), although all work was completed remotely.

\paragraph{G. SPEECH SITUATION}
All entries in our dataset were originally posted to Twitter and then collected by \citet{davidson2017automated}, who do not share when the posts were made.

\paragraph{H. TEXT CHARACTERISTICS}
All entries in our dataset are individual Twitter text posts, with a length of 140 characters or less.
We perform only minimal text cleaning, replacing user mentions (e.g., "@Obama") with "[USER]" and URLs with "[URL]".

\paragraph{I. LICENSE}
\citet{davidson2017automated} make the Twitter data they collected available for further research use via GitHub under an MIT license.\footnote{\href{https://github.com/t-davidson/hate-speech-and-offensive-language}{https://github.com/t-davidson/hate-speech-and-offensive-language}}
Our re-annotated subset of the data is made available under CC0-1.0 license at \href{https://github.com/paul-rottger/annotation-paradigms}{github.com/paul-rottger/annotation-paradigms}, so that the results of our experiment can be reproduced.

\paragraph{J. ETHICS APPROVAL}
We received approval for our experiment and the data annotation it entailed from our institution's ethics review board.\\

\section{Annotation Prompts} \label{app: prompts}
The three groups of annotators in our experiment all annotated the same data in the same order, but each group received different annotation prompts.
The full annotation guidelines for \textbf{G2} are available at \href{https://github.com/paul-rottger/annotation-paradigms}{github.com/paul-rottger/annotation-paradigms}.

\paragraph{G1 - Descriptive Group}
``Imagine you come across the post below on social media.
\textbf{Do you personally feel this post is hateful?}
We want to understand your own opinions, so try to disregard any impressions you might have about whether other people would find it hateful.''

\paragraph{G2 - Prescriptive Group}
``Imagine you come across the post below on social media.
\textbf{Does this post meet the criteria for hate speech?}
We are trying to collect objective judgments, so try to disregard any feelings you might have about whether you personally find it hateful.

Click here to view the criteria: LINK''

\paragraph{G3 - Control Group}
``Imagine you come across the post below on social media.
\textbf{Does this post meet the criteria for hate speech?}
A post is considered hate speech if it is 1) abusive and 2) targeted against a protected group (e.g., women) or at its members for being a part of that group.''

\end{document}